\documentclass[conference]{IEEEtran}
\IEEEoverridecommandlockouts
\usepackage{cite}
\usepackage{amsmath,amssymb,amsfonts}
\usepackage{algorithmic}
\usepackage{algorithm}
\usepackage{graphicx}
\usepackage{textcomp}
\usepackage{xcolor}
\usepackage{booktabs}
\usepackage{multirow}
\usepackage{threeparttable}
\usepackage{caption}
\def\BibTeX{{\rm B\kern-.05em{\sc i\kern-.025em b}\kern-.08em
    T\kern-.1667em\lower.7ex\hbox{E}\kern-.125emX}}
    
\begin{document}

\title{TwinLoop: Simulation-in-the-Loop Digital Twins for Online Multi-Agent Reinforcement Learning}

\author{%
  \IEEEauthorblockN{
  Nan Zhang\IEEEauthorrefmark{2}\IEEEauthorrefmark{1},
  Zishuo Wang\IEEEauthorrefmark{2},
  Shuyu Huang\IEEEauthorrefmark{2},
  Georgios Diamantopoulos\IEEEauthorrefmark{1}\IEEEauthorrefmark{2},
  \\ Nikos Tziritas\IEEEauthorrefmark{3},
  Panagiotis Oikonomou\IEEEauthorrefmark{3} 
  and
  Georgios Theodoropoulos\IEEEauthorrefmark{1}\IEEEauthorrefmark{2}
  \vspace{0.15cm}
  \IEEEauthorblockA{\IEEEauthorrefmark{1} Research Institute of Trustworthy Autonomous Systems, Southern University of Science and Technology, Shenzhen, China}
  \IEEEauthorblockA{\IEEEauthorrefmark{2} Department of Computer Science and Engineering, Southern University of Science and Technology, Shenzhen, China }
  \IEEEauthorblockA{\IEEEauthorrefmark{3} Department of Informatics and Telecommunications, University of Thessaly, Lamia, Greece}
  \thanks{This work was supported by the Research Institute of Trustworthy Autonomous Systems, the Guangdong Province Innovative and Entrepreneurial Team Programme (No. 2017ZT07X386) and  by the MSc programme in Informatics and Computational Biomedicine of the University of Thessaly. N. Zhang, G. Diamantopoulos, and G. Theodoropoulos are the corresponding authors.}
  }
}

\maketitle

\begin{abstract}
Decentralised online learning enables runtime adaptation in cyber-physical multi-agent systems, but when operating conditions change, learned policies often require substantial trial-and-error interaction before recovering performance. To address this, we propose \textit{TwinLoop}, a simulation-in-the-loop digital twin framework for online multi-agent reinforcement learning. When a context shift occurs, the digital twin is triggered to reconstruct the current system state, initialise from the latest agent policies, and perform accelerated policy improvement with simulation what-if analysis before synchronising updated parameters back to the agents in the physical system. We evaluate \textit{TwinLoop} in a vehicular edge computing task-offloading scenario with changing workload and infrastructure conditions. The results suggest that digital twins can improve post-shift adaptation efficiency and reduce reliance on costly online trial-and-error.
\end{abstract}


\section{Introduction}

Decentralised self-adaptation is increasingly important in cyber-physical multi-agent systems, where agents must continuously adapt their decisions to dynamically changing environments \cite{quin_decentralized_2021, muccini_self-adaptation_2016}. In such systems, online learning provides a natural basis for runtime adaptation, and a growing body of work has explored learning-based self-adaptation to cope with uncertainty and evolving operating conditions  \cite{gheibi_applying_2021, Cardellini2018, dangelo_decentralized_2020, dragan_coordinated_2025}. However, in highly dynamic environments, pure online learning approaches often suffer from costly exploration and slow convergence \cite{Cardellini2018, metzger_realizing_2024}. This is exacerbated in multi-agent scenarios where evolving agent behaviour becomes an additional unpredictable factor that affects environment dynamics \cite{dragan_coordinated_2025}.

A key challenge is that as the environment evolves, agent policies are increasingly likely to face conditions for which they do not possess the knowledge to handle \cite{padakandla_survey_2021}. To recover performance in these cases, agents typically need to adapt through trial-and-error interactions with the physical environment (online learning) \cite{lee_digital_2025}. However, such real-world exploration can be costly, as it can temporarily degrade system performance and, in safety- or resource-sensitive settings, even impose operational risks on the underlying infrastructure \cite{metzger_realizing_2024}.  A mechanism is therefore needed to rehearse adaptation before costly trial-and-error fully unfolds in the physical system.

Digital Twins (DTs) offer a promising foundation for such a mechanism. A DT maintains a synchronised virtual replica of the physical system, enabling state monitoring, environment reconstruction, and simulation-based what-if exploration~\cite{zhang2020}. 
Although prior studies have explored the use of DTs for reinforcement learning (RL) and optimisation~\cite{wang_digital_2022, zheng_digital_2024, diamantopoulos2022dynamic}, an important question remains insufficiently studied: can a DT, when triggered at runtime, use what-if simulation to accelerate the adaptation of decentralised online learning agents after environmental change?

Task offloading in Vehicular Edge Computing (VEC) provides a representative testbed for studying this problem \cite{uddin_intelligent_2025}. In VEC, vehicles continuously generate computation-intensive tasks, such as video processing and voice assistant services \cite{chen2025mcmfdrl}. Since each vehicle has limited onboard computational capacity, tasks can either be executed locally or offloaded to nearby Road-Side Units (RSUs) with stronger processing capability. Each vehicle is associated with a local decision agent that dynamically determines where tasks should be executed in order to minimise task completion latency \cite{chen2025mcmfdrl}. This setting is inherently dynamic due to temporally varying network conditions, changing topologies, vehicle mobility, and fluctuating task demand \cite{uddin_intelligent_2025}.

In this paper, we propose \textit{TwinLoop}, a DT-supported online learning framework that accelerates the adaptation of decentralised online learning agents. The crux of the system is a digital twin that mirrors the operating condition of the VEC system including agent policies. \textit{TwinLoop} leverages faster what-if simulation to enable policy exploration and adaptation to take place in the digital twin, with adapted policies then synchronised back to the physical system. The benefit of this approach is twofold: cost-free exploration and broader scenario coverage for better generalisation.

The main contributions of this paper are as follows:
\begin{itemize}
    \item We highlight the problem of costly adaptation in decentralised online learning after environmental shifts and motivate digital twins as a runtime what-if simulation mechanism for policy rehearsal.
    \item We propose \textit{TwinLoop}, a simulation-in-the-loop DT framework in which a runtime-triggered twin reconstructs the current system state, performs accelerated policy improvement in simulation, and synchronises updated parameters back to the physical system.
    \item We evaluate \textit{TwinLoop} in a VEC task-offloading scenario and show that it improves post-shift adaptation efficiency and reduces reliance on costly real-world trial-and-error.
\end{itemize}

\section{Related Work}

\subsection{Online Decentralised Learning in Dynamic Environments}

Existing work on decentralised self-adaptation through online learning mainly addresses trial-and-error coordination in real-world environments. Cardellini et al.~\cite{Cardellini2018} adopt a hierarchical design in which local learners make adaptation decisions and a central coordinator resolves conflicts, whereas D'Angelo et al.~\cite{dangelo_decentralized_2020} advocate a fully decentralised scheme based on inter-agent information sharing and trust-aware updates. Dragan et al.~\cite{dragan_coordinated_2025} coordinate multiple learners through factored Q-functions over local and shared goals. 
However, these studies still rely on costly exploration in the real environment rather than pre-adaptation through simulation.

\subsection{Digital Twins for Online Learning}

Digital twins have increasingly been explored as a means to support online learning. In early work, we explored simulation for prescriptive what-if analysis and data-driven adaptation \cite{kennedy_intelligent_2006, theodoropoulos_dddas_2023}; more recently, we have investigated knowledge management and learning in cognitive digital twins \cite{10.1145/3635306,10.1007/978-3-031-94895-4_8,10.1007/978-3-031-52670-1_23,10.1007/978-981-95-8411-6_39}.

Most prior DT-assisted online learning studies focus on \emph{single-agent} settings. Lee et al.~\cite{lee_digital_2025} use a DT to temporarily train a decision policy at runtime decision points before applying it to the real system. Wang et al.~\cite{wang_digital_2022} let the agent execute one action in reality while exploring additional actions in the twin, and combine real and simulated rewards for policy improvement. Wu et al.~\cite{wu_digital_2021} establish a continuous feedback loop in which real trajectories improve the DT, while the DT in turn supports reinforcement learning. Deng et al.~\cite{deng_digital_2021} enable dynamic switching between expert knowledge and an online reinforcement learning policy. 

In distributed and multi-agent learning, Sun et al.~\cite{sun_adaptive_2021} use a descriptive DT to monitor IoT node status and adapt federated aggregation frequency. Overall, prior work either uses DTs for single-agent simulation-in-the-loop learning or adopts descriptive DTs for coordination support, with limited attention to continuously updated DTs for online simulation in decentralised multi-agent learning.

\subsection{Digital Twin-Assisted Decentralised and Multi-Agent Learning for Task Offloading in VEC}

In VEC task offloading, DT-assisted learning remains limited. Following the taxonomy introduced in our earlier work~\cite{zhang2020}, existing studies are still largely confined to \emph{descriptive} twins for current state mirroring and \emph{predictive} twins for near-future estimation, while \emph{prescriptive} DTs based on online what-if simulation-in-the-loop analysis remain largely unexplored. Most decentralised and multi-agent studies are descriptive. Zhang et al.~\cite{zhang_adaptive_2022} mirror network and resource states to estimate cooperation gains and enable adaptive agents aggregation; while some other works focus on mirroring per-vehicle task processing context~\cite{xie_digital_2024} and vehicle/RSU states~\cite{singh_genai_2025}. By contrast, predictive DTs further estimate future task arrivals and throughput to support offloading and resource reservation~\cite{zheng_digital_2024}.

\section{System Model and Problem Formulation}

\begin{figure}
    \centering
    \includegraphics[width=1\linewidth]{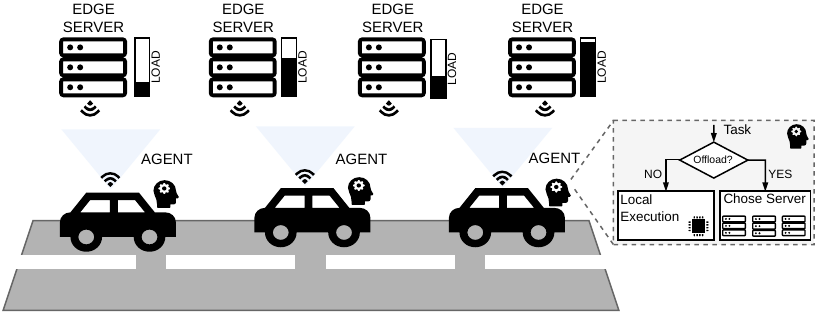}
    \caption{System model under consideration.}
    \label{fig:placeholder}
\end{figure}

We consider a VEC scenario with $M$ vehicles and $N$ RSU edge servers as illustrated in Fig.~\ref{fig:placeholder}. Each vehicle generates computation tasks following a Poisson process with arrival rate $\lambda$. Upon task arrival, each vehicle independently decides whether to process the task locally or offload it to one of the servers via a Vehicle-to-Infrastructure (V2I) wireless link, with the objective of minimising end-to-end completion latency.

 \begin{figure*}
    \centering
    \includegraphics[width=0.9\linewidth]{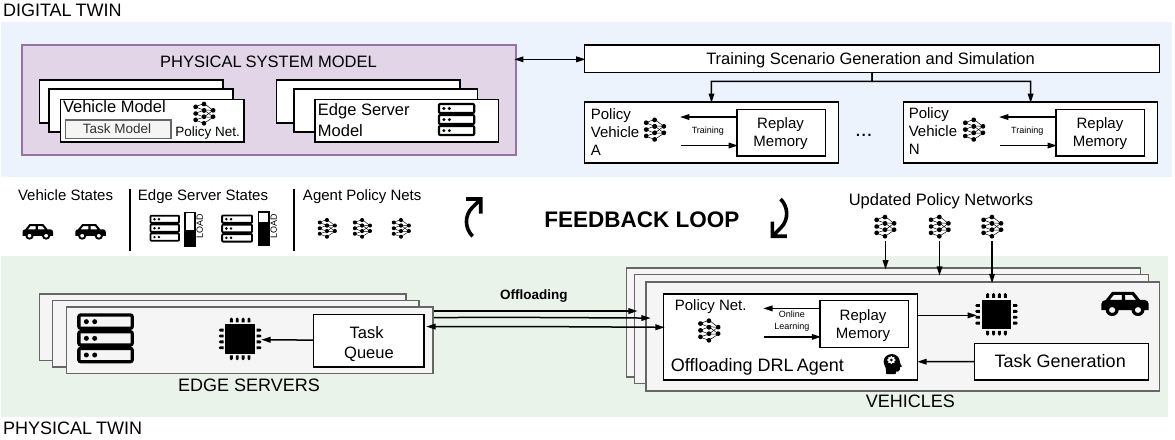} 
    \caption{Architecture of the DT-assisted VEC system.}
    \label{fig:arch}
\end{figure*}

\subsection{System Model}

\noindent\textbf{Communication Model.}
Following~\cite{chen2025mcmfdrl}, the uplink transmission rate between vehicle $v_i$ and server $s_j$ is modelled as:

\begin{equation}
  r_{ij} = B \log_2\!\left(1 + \frac{P \cdot d_{ij}^{-\alpha}}{\sigma^2}\right)
  \label{eq:rate}
\end{equation}
where $B$ is the V2I channel bandwidth, $P$ is the vehicle uplink transmit power, $d_{ij}$ is the Euclidean distance between $v_i$ and $s_j$, $\alpha$ denotes the path-loss exponent, and $\sigma^2$ represents the background noise power. The transmission delay for a task with input data size $b_i$ is (assuming negligible downlink delay):
\begin{equation}
  t_{ij}^{trans} = \frac{b_i}{r_{ij}}
  \label{eq:trans_delay}
\end{equation}
 
\noindent\textbf{Computation Model.}
Each server $s_j$ maintains a processing rate $f_j$ and a cumulative backlog $L_j$ of pending computation cycles from previously accepted tasks. The computation delay for a task with demand $D_i$ at server $s_j$ is modelled as:
 
\begin{equation}
  t_{ij}^{comp} = \frac{L_j + D_i}{f_j}
  \label{eq:comp_server}
\end{equation}
 
The local computational platform of a vehicle is similarly modelled using a local processing rate $f_i$ and its cumulative cycle backlog $L_i$. The computation delay for processing a task with demand $D_i$ locally on vehicle $i$ is:

\begin{equation}
  t_i^{local} = \frac{L_i + D_i}{f_i}
  \label{eq:comp_local}
\end{equation}
 
\noindent\textbf{End-to-End Latency.}
The end-to-end task completion latency is:
 
\begin{equation}
  t_i^{e2e} =
  \begin{cases}
    t_i^{local},                  & \text{local execution} \\
    t_{ij}^{trans} + t_{ij}^{comp}, & \text{offloaded to $s_j$}
  \end{cases}
  \label{eq:e2e}
\end{equation}

\noindent\textbf{Objective:} The objective of each vehicle $i$ is to minimise the total end-to-end latency of all tasks:
 
\begin{equation}
  \min\sum_{k=1}^{K_i} t_{i,k}^{e2e}
  \label{eq:objective}
\end{equation}
where $K_i$ is the total number of tasks generated by vehicle $i$.
\subsection{Reinforcement Learning Formulation}
 
We model the task offloading decision of each vehicle as a Markov Decision Process (MDP), defined by the tuple $\langle \mathcal{S}, \mathcal{A}, \mathcal{S}', \mathcal{R} \rangle$, where $\mathcal{S}'$ denotes the next state space reached after executing an action. Each vehicle maintains an independent agent that observes local information and selects an offloading target without coordination with other vehicles.

\noindent\textbf{State Space.}
At each decision step, vehicle $v_i$ constructs a local observation vector:
\begin{equation}
  s_i = \bigl(f_i,\; L_i,\; b_i,\; D_i,\;\{f_j,\, L_j,\, r_{ij}\}_{j=1}^{N}\bigr)\label{eq:state}
\end{equation}
where $f_i$, $L_i$ are the vehicle's local processing rate and backlog; $b_i$, $D_i$ are the task data size and computation demand; and $f_j$, $L_j$, $r_{ij}$ describe the processing rate, backlog, and uplink rate of each server $s_j$.
 
\noindent\textbf{Action Space.}
The action $a_i \in \{0, 1, \ldots, N\}$ selects the execution target. $a_i = 0$ denotes local execution; $a_i = j > 0$ denotes offloading to server $s_{j}$.
 
\noindent\textbf{Reward Function.}
The reward is the negative end-to-end latency of the completed task:
\begin{equation}
  r_i = -\,t_i^{e2e}\label{eq:reward}
\end{equation}
Maximising the cumulative discounted reward therefore directly minimises the objective in~\eqref{eq:objective}.
 
\noindent\textbf{Learning Algorithm.}
Each vehicle agent is trained independently using a Dueling Double DQN (D3QN)~\cite{wang2016dueling}. Unlike standard DQN, Dueling DQN decouples the estimation of the state-value function $V(s; \theta_V)$ and the state-dependent advantage function $A(s, a; \theta_A)$. The Q-value is given by:
\begin{equation*}
    Q(s, a; \theta) = V(s; \theta_V) + \left( A(s, a; \theta_A) - \frac{1}{|\mathcal{A}|} \sum_{a'} A(s, a'; \theta_A) \right)
\end{equation*}
where $\theta = \{\theta_V, \theta_A\}$ denotes the network parameters of a single agent. The Double DQN component mitigates Q-value overestimation by decoupling action selection from action evaluation using separate online and target networks.

\noindent\textbf{Exploration Policy.}
Action selection follows a Boltzmann policy. Given the Q-values $Q(s, a; \theta)$ for valid actions, the probability of selecting action $a$ is:
\begin{equation}
    \pi(a \mid s) = \frac{\exp\bigl(Q(s,a;\theta) / \tau\bigr)}{\sum_{a'} \exp\bigl(Q(s,a';\theta) / \tau\bigr)}
\end{equation}
where $\tau > 0$ is a temperature parameter; a higher $\tau$ encourages exploration, while $\tau \to 0$ leads to a greedy policy.

\subsection{DT-Assisted Adaptive Offloading}

The proposed framework consists of a Digital Twin (DT) and its corresponding Physical Twin (PT), where the PT denotes the physical system being mirrored and assisted by the DT, as illustrated in Fig.~\ref{fig:arch}. The framework operates in three stages:

\begin{figure}
    \centering
    \includegraphics[width=\linewidth]{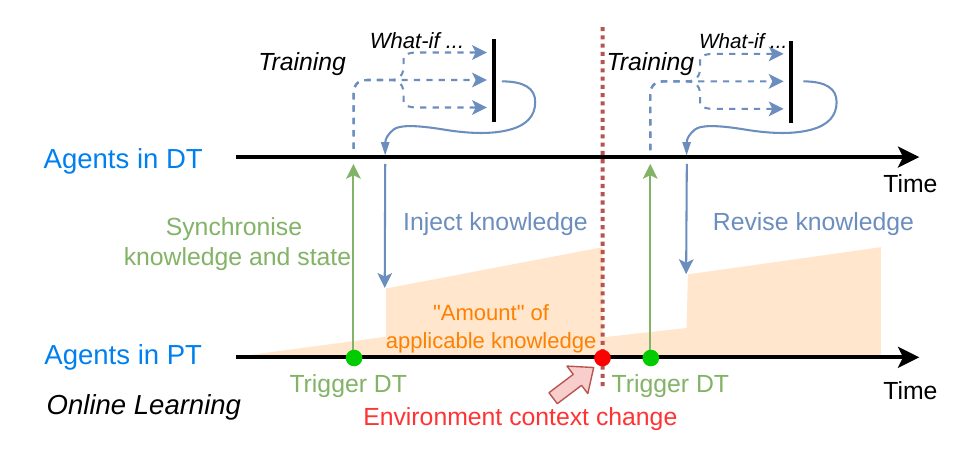} 
    \caption{Workflow of the DT-assisted VEC system.}
    \label{fig:workflow}
\end{figure}

\begin{algorithm}
\caption{DT-Assisted Adaptation}
\label{alg:dt}
\begin{algorithmic}[1]

\REQUIRE DT training budget $T_{\text{DT}}$
\ENSURE Updated agent weights $\Theta_{\text{PT}}$ in the PT

\STATE \textbf{Stage 1: Snapshot} 
\STATE \quad Acquire the current PT system snapshot $\Phi$

\medskip
\STATE \textbf{Stage 2: DT Training}
\STATE \quad Reconstruct and initialise the DT from $\Phi$
\STATE \quad Set exploration temperature $\tau \leftarrow \tau_0$ 
\FOR{each step during $T_{\text{DT}}$}
    \STATE Observe $s$; select $a$ via Boltzmann policy
    \STATE Execute $a$; observe $r = -t^{e2e}$ and $s'$
    \STATE Store $(s, a, r, s')$ in replay buffer $\mathcal{B}$
    \STATE Update $\Theta_{\text{DT}}$ using D3QN
\ENDFOR

\medskip
\STATE \textbf{Stage 3: Weight Synchronisation}
\STATE \quad Synchronise weights back to the PT: $\Theta_{\text{PT}} \leftarrow \Theta_{\text{DT}}$
\STATE \quad Resume online learning in the PT 
\end{algorithmic}
\end{algorithm}

\noindent{\textbf{\textit{Stage 1: Snapshot.}}}
Upon triggering, the DT acquires a snapshot $\Phi$ of the current PT system state:
\begin{equation}
  \Phi = \bigl(\{f_j, L_j, p_j\}_{j=1}^{N},\;\{f_i, L_i, p_i\}_{i=1}^{M},\;
               \lambda,\; \mathbf{w},\; \Theta_{\text{PT}}\bigr)
  \label{eq:snapshot}
\end{equation}
where $\{f_j, L_j, p_j\}$ are the servers' processing rates, backlogs, and positions, $\{f_i, L_i, p_i\}$ are the vehicles' processing rates, backlogs, and positions, $\lambda$ is the task inter-arrival rate, $\mathbf{w}$ is the task-type distribution, and $\Theta_{\text{PT}} = \{\theta_i\}_{i=1}^{M}$ is the current set of network weights of all agents. 
In a real distributed system, obtaining a globally consistent snapshot $\Phi$ can be challenging as it requires synchronisation across multiple agents. In this work, we assume that such snapshots are available as their acquisition falls outside the scope of this paper.

\noindent{\textbf{\textit{Stage 2: DT Training.}}}
The DT uses $\Phi$ to initialise both the simulated environment and the agent network weights. The DT starts with a high exploration rate ($\tau = \tau_0$) at the beginning of training, allowing agents to rapidly discover strategies suited to the new environment conditions through simulation. 
 
\noindent{\textbf{\textit{Stage 3: Weight Synchronisation.}}}
Upon completion of DT training, the updated weights $\Theta_{\text{DT}}$ are synchronised back to the PT. Each vehicle then updates its local agent with $\Theta_{\text{DT}}$ and continues online learning in the physical environment. The above procedure is formally described in Algorithm~\ref{alg:dt} and the workflow of the proposed approach is illustrated in Fig.~\ref{fig:workflow}.

\section{Experimental Setup}

\subsection{Simulation Settings}
\begin{figure}[t]
    \centering
    \includegraphics[width=0.8\linewidth]{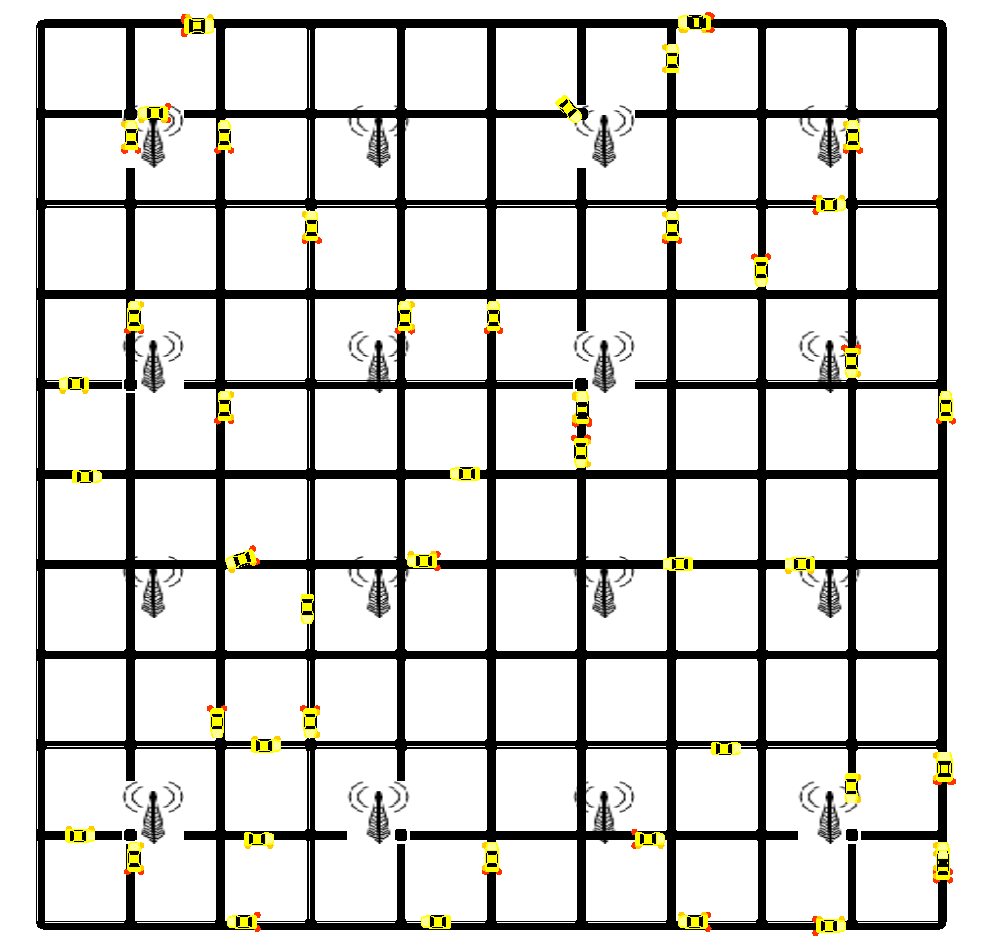} 
    \vspace{-2mm}
    \caption{2km $\times$ 2km simulation scenario.}
    \label{fig:sumo-gui}
    \vspace{-5mm}
\end{figure}

\begin{figure*}[!ht]
\centering

\begin{minipage}{\textwidth}
\begin{threeparttable}
    \captionof{table}{Phase-wise end-to-end latency (seconds) for all methods.}
    \label{tab:full_results}

    \setlength{\tabcolsep}{4.5pt}
    \renewcommand{\arraystretch}{1.05}
    \footnotesize
\begin{tabular}{ll r@{\;}r@{\;}r r@{\;}r@{\;}r r@{\;}r@{\;}r r@{\;}r@{\;}r}
\toprule
& & \multicolumn{3}{c}{Warmup (0--500\,s)} 
  & \multicolumn{3}{c}{Phase~1 (500--1500\,s)}
  & \multicolumn{3}{c}{Phase~2 (1500--2500\,s)}
  & \multicolumn{3}{c}{Phase~3 (2500--3500\,s)} \\
\cmidrule(lr){3-5}\cmidrule(lr){6-8}\cmidrule(lr){9-11}\cmidrule(lr){12-14}
Category & Method
  & Mean & P90 & P99
  & Mean & P90 & P99
  & Mean & P90 & P99
  & Mean & P90 & P99 \\
\midrule
\multirow{3}{*}{Baselines}
  & Random
    & 7.404 & 15.180 & 37.575
    & 113.712 & 349.580 & 606.062
    & 1141.270 & 3324.453 & 4571.616
    & 558.378 & 1576.397 & 1793.818 \\
  & Online
    & 6.731 & 13.065 & 30.003
    & 18.293 & 43.060 & 82.834
    & 11.856 & 19.785 & 66.930
    & \textbf{4.240} & \textbf{7.708} & 18.031 \\
  & Offline
    & \textbf{4.282} & \textbf{7.045} & \textbf{13.793}
    & 27.031 & 55.224 & 95.262
    & 36.974 & 74.408 & 133.062
    & 5.741 & 10.168 & 49.572 \\
\midrule
\multirow{2}{*}{\shortstack[l]{Exploit}}
  & Exploit ($T_\text{DT}$=250)
    & 6.342 & 12.896 & 24.270
    & 11.626 & 22.922 & 44.726
    & 14.020 & 22.594 & 66.986
    & 4.395 & 7.770 & \textbf{17.228} \\
  & Exploit ($T_\text{DT}$=500)
    & 6.944 & 13.335 & 36.308
    & 9.678 & 17.431 & 36.053
    & 10.979 & 18.629 & 35.848
    & 4.567 & 8.496 & 18.340 \\
\midrule
\multirow{3}{*}{DT Single-Scenario}
  & $k$=1, $T_\text{DT}$=500
    & 6.689 & 13.075 & 27.602
    & 10.187 & 19.396 & 39.223
    & \textbf{10.529} & \textbf{16.761} & \textbf{29.889}
    & 4.532 & 8.570 & 18.134 \\[1pt]
  & $k$=2, $T_\text{DT}$=500
    & 6.338 & 12.645 & 23.575
    & \textbf{8.320} & 15.071 & 30.482
    & 10.698 & 17.372 & 34.961
    & 4.380 & 7.905 & 18.619 \\[1pt]
  & $k$=4, $T_\text{DT}$=500
    & 6.699 & 13.848 & 24.999
    & 9.676 & 16.705 & 34.632
    & 12.151 & 20.517 & 45.506
    & 4.683 & 8.600 & 18.753 \\
\midrule
\multirow{2}{*}{DT Multi-Scenario}
  & $k$=1, $T_\text{DT}$=500
    & 7.067 & 14.326 & 30.676
    & 9.828 & 17.541 & 36.990
    & 11.386 & 18.795 & 38.182
    & 4.859 & 8.808 & 21.141 \\[1pt]
  & $k$=2, $T_\text{DT}$=500
    & 6.508 & 12.980 & 24.445
    & 8.448 & \textbf{14.238} & \textbf{27.291}
    & 12.986 & 21.114 & 44.444
    & 4.523 & 8.433 & 18.401 \\
\bottomrule
\end{tabular}
\begin{tablenotes}
    \item \textit{Note:} Each phase covers 1,000\,s of simulation time. \textbf{Bold} denotes the best value in each column.
    \item $k$: number of DT triggers per phase; $T_\text{DT}$: scenario duration; Multi-Scenario: 3 scenarios per trigger.
\end{tablenotes}
\end{threeparttable}
\end{minipage}

\begin{minipage}[b]{0.49\textwidth}
    \centering
    \includegraphics[width=\linewidth]{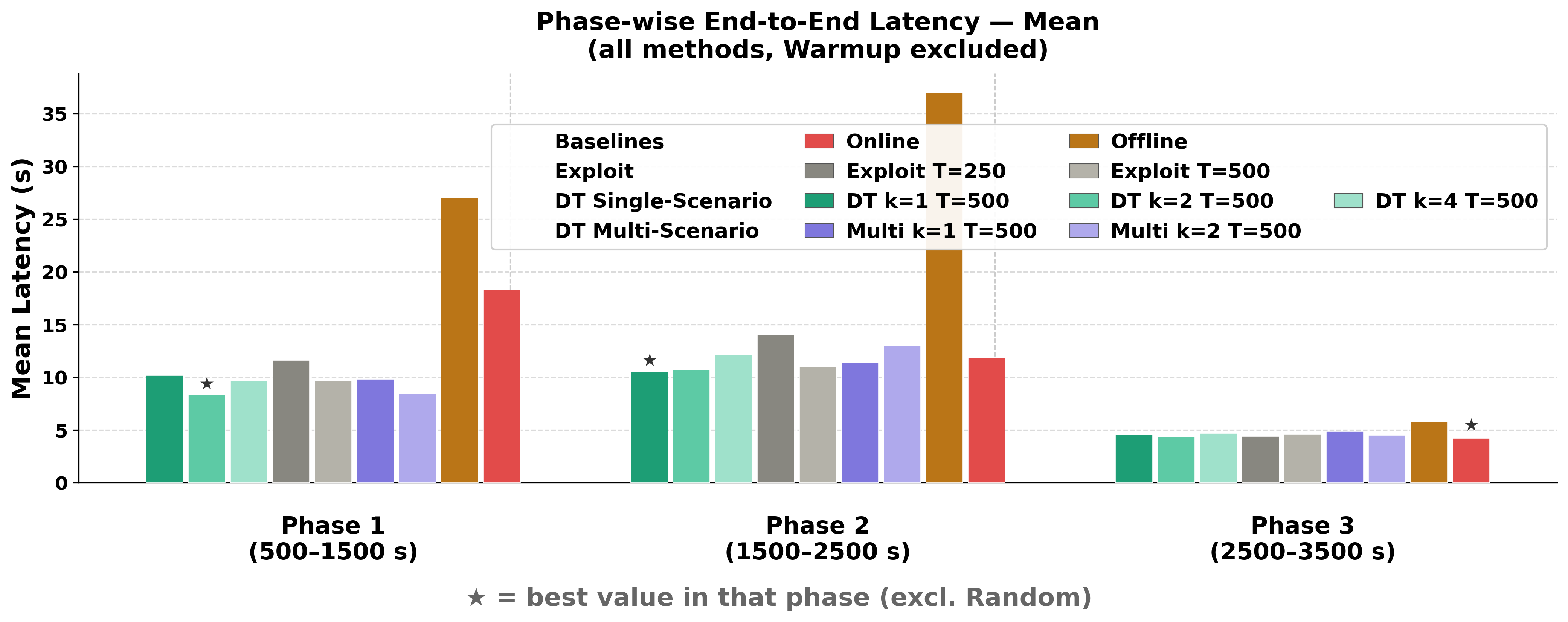}
    \caption{Phase-wise mean latency across methods.}
    \label{fig:bar_chart}
\end{minipage}
\hfill
\begin{minipage}[b]{0.49\textwidth}
    \centering
    \includegraphics[width=\linewidth]{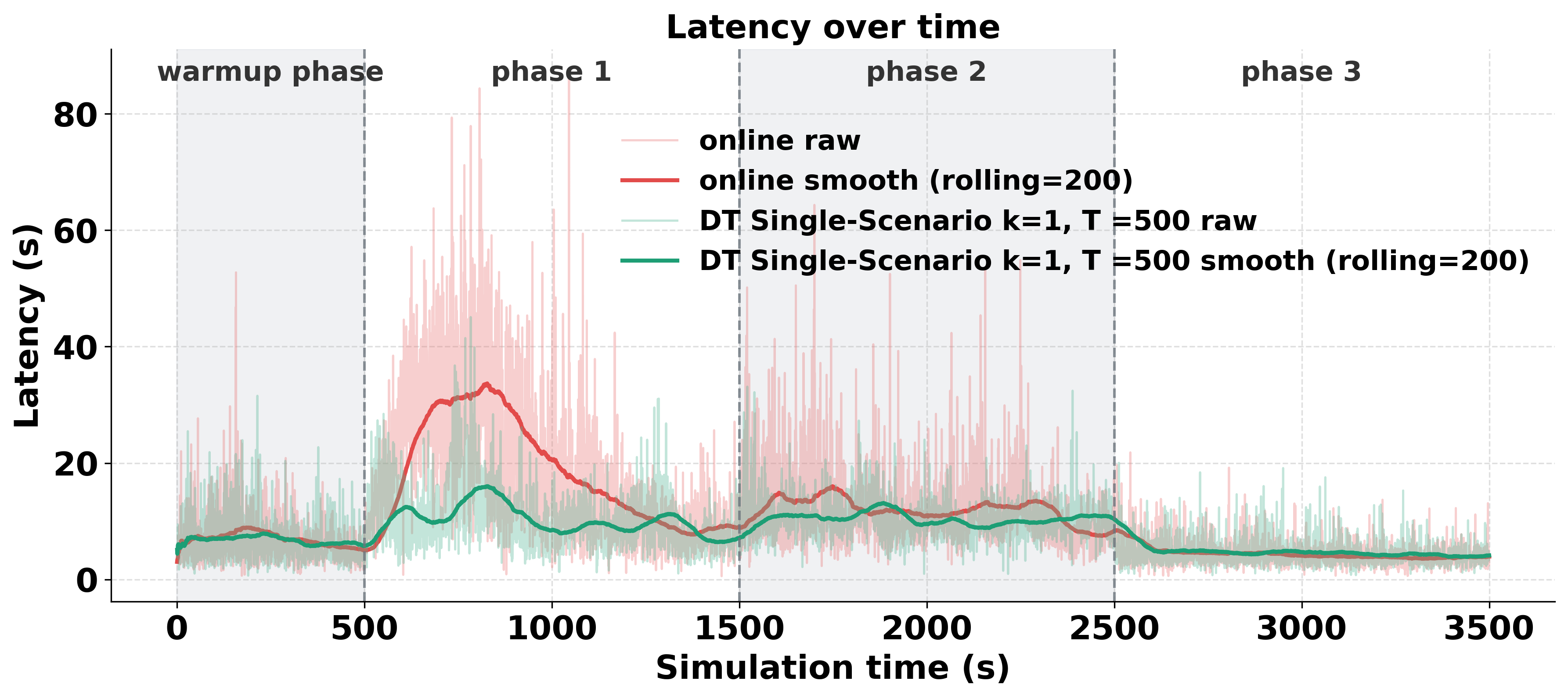}
    \caption{Latency trajectory: Online vs. DT-assisted adaptation.}
    \label{fig:dt_online}
\end{minipage}

\vspace{-5mm}

\end{figure*}

Experiments are conducted on a fixed 2\,km\,$\times$\,2\,km grid road network simulated using SUMO with $N = 16$ stationary RSU edge servers. To evaluate adaptation to dynamic conditions, 
we use a 3\,500\,s simulation. It includes a 500\,s warm-up phase (excluded from comparative analysis), followed by three 1\,000\,s phases, each introducing an environmental change.

The initial system configuration in the warm-up phase follows the setup in~\cite{chen2025mcmfdrl}. The wireless channel adopts a path-loss model ($P = 30$\,dBm, $\alpha = 4.0$, $B = 1$\,MHz, $\sigma^2 = 10^{-13}$\,W). The 16 RSU servers comprise four low-tier (2.0\,GHz), four mid-tier (2.5\,GHz), and eight high-tier (3.5\,GHz) nodes. We deploy 45 active vehicles with onboard processing capacities drawn uniformly from $[0.8, 1.2]$\,GHz. Task arrivals follow a Poisson process with rate $\lambda = 1/8\,\mathrm{s}^{-1}$. Six task types are considered, with computation demands ranging from $10^9$ to $10^{10}$\,cycles and data sizes of $5\!\times\!10^5$--$10^7$\,bits.

Each subsequent phase modifies the initial configuration as follows:
\textit{Phase~1} (500--1\,500\,s): Active vehicles increase to 65, and workloads shift toward computation-heavy tasks (${\sim}10^{10}$\,cycles).
\textit{Phase~2} (1\,500--2\,500\,s): Half of the servers enter maintenance, degrading processing rates by 50--70\%, while the task arrival rate increases by 20\%.
\textit{Phase~3} (2\,500--3\,500\,s): All server capacities are restored, but the dominant workload shifts to lightweight tasks.

These phases aim to emulate realistic traffic and infrastructure fluctuations seen in urban IoV systems. The increase in vehicles and computation-heavy workloads (Phase 1) mirrors rush-hour congestion where more users generate latency-sensitive applications, while partial server degradation reflects maintenance windows or unexpected outages in edge infrastructure. The final shift to lightweight tasks (Phase 3) captures off-peak conditions where demand stabilises and applications become less resource-intensive.

\subsection{Experiment Settings}
\label{sec:exp_settings}
To verify the effectiveness of the DT in accelerating policy convergence, we compare the proposed DT-assisted approach against the following baselines: \textbf{(a) Random}: assigns tasks uniformly across all servers and local execution; \textbf{(b) Online}: performs online learning exclusively in the PT without DT assistance; \textbf{(c) Offline}: deploys the converged weights obtained from the full \textit{Online} run; and \textbf{(d) Exploit}: operates identically to the proposed method, but switches the agents in PT to exploitation ($\tau = \tau_{\min}$) after weight synchronisation. 
Both the PT and DT use SUMO\footnote{https://eclipse.dev/sumo/} for vehicle mobility simulation, with wireless channel, task offloading, and RL logic implemented in Python. The DT runs $\sim$25$\times$ faster than the PT. Experiments\footnote{Code, data, and experiment scripts are available at: https://github.com/asia-lab-sustech/TwinLoop} were run on an AMD Ryzen 7 6800H CPU with 16\,GB RAM and an NVIDIA RTX 3060 GPU under Ubuntu 24.04.

To investigate the factors affecting DT effectiveness, we evaluate variants along three dimensions: DT triggers per phase ($k \in \{1, 2, 4\}$), DT training time per trigger ($T_{\text{DT}} \in \{250, 500\}$), and single-scenario versus multi-scenario simulation. 
All scenarios are initialised from the received PT snapshot, with vehicle routes randomly assigned in each case. In the multi-scenario setting, the DT runs a sequence of perturbed scenarios, carrying model weights over between them. Perturbations ($5\%$) are applied to parameters such as vehicle speed, task computation demand, and task data size.

\section{Experimental Results}

We evaluate the proposed DT-assisted framework against the baselines and variants defined in Section~\ref{sec:exp_settings}. Performance is evaluated using the per-task end-to-end latency $t^{e2e}$. We report the phase-wise mean latency, 90th-percentile (P90), and 99th-percentile (P99). Table~\ref{tab:full_results} summarises the phase-wise mean, P90, and P99 latencies while Fig.~\ref{fig:bar_chart} provides a visual overview of the phase-wise mean latency across all methods.

The results demonstrate that the DT functions as a convergence accelerator. DT-assisted training substantially reduces latency following phase transitions, with the benefit concentrated in the initial period following each environmental change (Fig.~\ref{fig:dt_online}). Compared to the \textit{Online} baseline, the proposed method (DT single-scenario, $k=1$, $T_\text{DT}=500$s) yields significant initial gains in Phase~1. In Phase~2, the mean gap narrows yet the P99 reduction remains significant. The \textit{Offline} baseline shows severe performance degradation across phases. This is consistent with the premise that offline learning methods struggle in highly dynamic environments.

In Phase~3 all methods, excluding random, exhibit a similar latency profile. This is primarily attributed to the operating conditions of phase 3, decreased task computational requirements and recovered edge server capacity, that model off-peak settings. This decreases the pressure put on the agents allowing all approaches to achieve low latency. 

Focusing the comparison on the duration of training in the DT, the \textit{Exploit} baseline is used to evaluate the effects of training for $T_\text{DT}=250$s and $T_\text{DT}=500$s. Notably, \textit{Exploit} is chosen to eliminate the noise added by exploration in the PT. The results of the analysis show that the longer trained variant achieves a substantial recrudesce in latency. This highlights the fact that longer training times are beneficial as they allow the policy to better adapt to the new conditions.

Focusing on the analysis on the DT trigger frequency ($k$) results show that triggering DT adaptation can enhance performance under high environmental pressure but too frequent updates have destabilising effects. Specifically, results show that increasing $k$ from 1 to 2 leads to a reduction in latency under Phase~1, but a further increase to $k=4$ leads to a regression in performance under all phases.

Focusing on multi-scenario vs. single-scenario DT-assisted adaptation, the differences are marginal. This can be primarily attributed to the fact that the simulation scenario used for evaluation models a homogeneous urban setting diminishing the value of training on varying scenarios. We expect multi-scenario training to lead to more impactful results under heterogeneous scenarios covering both urban and rural settings; such evaluation constitutes future work. 

Overall, the results show that DT assistance is most effective during the most demanding periods (Phase 1 and 2) effectively reducing mean latency and suppressing tail values. The results further suggest that trigger frequency plays a critical role: triggering the DT too rarely limits its adaptation benefit, whereas overly frequent triggering can harm performance. This highlights adaptive DT trigger scheduling as an important direction for future work.

\section{Conclusion and Future Direction}

This paper proposed a Digital Twin-assisted deep reinforcement learning framework to accelerate online learning convergence in multi-agent environments. Using vehicular edge computing (VEC) task offloading as a representative case study, experimental results have demonstrated the ability of the proposed DT-assisted online policy adaptation to significantly accelerate policy convergence leading to decreased end-to-end task execution latency. Future work directions include the development of an adaptive DT triggering mechanism based on detecting context shifts and evolving conditions; a policy merging scheme to allow for policy augmentation rather than replacement, as well as conducting further evaluation of the framework in highly heterogeneous network topologies to explore its generalisation capabilities.

\bibliographystyle{IEEEtran}
\bibliography{ref.bib}

\end{document}